\documentclass[runningheads]{llncs}
\usepackage[T1]{fontenc}
\usepackage{graphicx}
\usepackage{textcomp}
\usepackage{eurosym}
\usepackage{booktabs}
\usepackage{amsmath,amssymb,amsfonts}
\usepackage{soul, xcolor}
\usepackage{float}
\usepackage{booktabs} 
\usepackage{appendix}

\begin{document}
\title{\textcolor{black}{Flow-aware Optimal Navigation in Unsteady Flows through Reinforcement Learning}}
\titlerunning{\textcolor{black}{Flow-aware Optimal Navigation in Unsteady Flows through RL}}
%
\author{Andrea Braghin\inst{1} \and
Nicolò Botteghi \inst{2} \and
Matteo Tomasetto \inst{1} \and \\
Andrea Manzoni \inst{2} \and
Gabriele Cazzulani \inst{1}}
\authorrunning{A. Braghin, N. Botteghi, M. Tomasetto, A. Manzoni, G. Cazzulani}
%
\institute{Department of Mechanical Engineering, Politecnico di Milano, Milano, Italy
\email{andreamaria.braghin@polimi.it}\\
\and
MOX, Department of Mathematics, Politecnico di Milano, Milano, Italy}
\maketitle              
\begin{abstract}
Autonomous robotic navigation in nonstationary time-varying fluid flows remains a fundamental challenge due to partial observability and the unpredictability of realistic environments. While classical optimal control frameworks employed in robotics require unrealistic a-priori global flow knowledge, biological systems are able to navigate successfully by exploiting localized sensory cues. In this work we present a reinforcement learning approach using the TD3 algorithm to train autonomous agents to reach arbitrary targets within a parametric, chaotic double-gyre flow. To investigate optimal sensory mechanisms, we evaluate five bio-inspired observation strategies based on relative position, local velocity or local vorticity measures, and short-term memory variants. Additionally, we analyze the impact of providing agents with explicit global flow parameters. Numerical results demonstrate that an agent that is able to sense and remember a set number of flow velocity measures achieves the highest performance. The experiments reveal a trade-off in sensor utility: velocity-aware agents optimize energy efficiency, whereas vorticity sensors provide superior structural mapping and achieve better target proximity. Incorporating explicit global flow parameters is shown to decrease navigation performance. This behavior suggests that reinforcement learning-based autonomous systems develop more robust and general policies when restricted to implicit flow representations. The presented results offer insights for improving the transition of bio-inspired robotic navigation from simulation to real-world environments.

\end{abstract}
\section{Introduction}
Efficient navigation in the presence of unsteady flow velocity fields represents a fundamental challenge for autonomous robotic systems \cite{gunnarson2021learning}. This challenge arises across a wide spectrum of applications, including ocean surveying \cite{zhang2008optimal}, monitoring of animal communities \cite{kuhnz2020benthic}, drone-based inspection in windy conditions \cite{guerrero2013uav}, surveillance and traffic monitoring in urban environments, and weather balloon station keeping \cite{bellemare2020autonomous}. 
In these scenarios, robots must operate under the influence of unsteady fluid phenomena such as wind gusts, turbulent wakes, and ocean currents while autonomously reaching target locations and performing their tasks. 
When the background velocity field is fully known a priori, optimal navigation can be formulated within well-established theoretical frameworks, ranging from the classical Zermelo navigation problem in optimal control theory \cite{zermelo1931navigationsproblem} to modern computational optimization approaches \cite{zhang2008optimal,kularatne2018going,krishna2022finite}. However, such assumptions rarely hold in practice. Real-world flow environments are typically time-varying, partially observable through onboard sensors, computationally intractable at full fidelity, and often characterized by chaotic dynamics that are difficult to predict. 
As autonomous systems increasingly operate without continuous external supervision, they must rely primarily on onboard sensing to interpret and respond to local flow conditions.

Biological systems demonstrate that effective navigation does not require full-field knowledge but can emerge from intelligent exploitation of local flow information. For example, fish use their lateral line system to detect velocity gradients and identify obstacles by sensing vorticity variations associated with boundary layers \cite{oteiza2017novel}. 
These observations suggest that flow-aware navigation based on structured local sensing offers a promising alternative to global optimal-control formulations in complex and uncertain environments \cite{gunnarson2021learning}.

Recent advances in machine learning have opened new opportunities for tackling complex control problems. In particular, reinforcement learning (RL) has emerged as a powerful framework for decision-making in high-dimensional and uncertain environments \cite{sutton_reinforcement_2018}.
Early studies have demonstrated the potential of this approach in flow-driven environments \cite{gunnarson2021learning,colabrese2017flow,gustavsson2017finding}.
Despite these promising results, RL often suffers from limited data efficiency, high training costs, and sensitivity to reward design. Eventually, simulation-to-real transfer becomes particularly challenging in high-dimensional, turbulent, and partially observed flow environments. Bridging this gap is essential for advancing RL from proof-of-concept studies toward reliable, flow-aware navigation in real-world environments.

In this work, we employ state-of-the-art RL for learning general navigation policies in unsteady flows described by a double-gyre equation. In particular, through extensive numerical investigation, we compare and analyze (\textit{i}) different (flow) observation strategies, e.g. velocity and/or vorticity, and the importance (\textit{ii}) of memory, and (\textit{iii}) of global parameters of the environment to understand which information is critical for efficient flow navigation in unsteady and chaotic environments.

\section{Background}
\label{background}
RL is a promising machine learning strategy to solve sequential decision-making problems through trial-and-error interactions. RL aims to design intelligent agents that take actions in an unknown environment, i.e., selects control inputs to maximize their cumulative reward~\cite{sutton_reinforcement_2018}. Instead of relying on a model of the underlying dynamics, RL derives optimal feedback control strategies from system measurements and task-dependent reward samples obtained through different interactions with the (possibly unknown) environment. This framework is suitable for a wide range of sequential decision-making problems, such as game playing, robotic control, autonomous driving, and resource management. 

Formally, RL problems are often modeled using Markov Decision Processes (MDPs), which provide a mathematical framework for modeling decision-making in situations where outcomes are partly random and partly under the control of the agent. A Markov decision process is defined by a tuple $(\mathcal{S}, \mathcal{A}, T, R)$, where $\mathcal{S} \subset \mathbb{R}^{N_s}$ and $\mathcal{A} \subset \mathbb{R}^{N_a}$ denote the state and action spaces, respectively, $T:\mathcal{S}\times \mathcal{S}\times \mathcal{A} \to [0,1]$ indicates the transition probability for any state $\mathbf{s}, \mathbf{s}' \in \mathcal{S}$ and action $\mathbf{a} \in \mathcal{A}$, and $R:\mathcal{S}\times \mathcal{A}\to \mathbb{R}$ indicates the reward function that determines the instantaneous reward the agent receives for each action taken in a given state.  

For a given time instant $t$, the agent observes the state of the environment $\mathbf{s}$ and takes action $\mathbf{a}$. In MDP settings, the state of the agent corresponds to the state of the environment. As a result, the environment changes its state to some $\mathbf{s}'$ according to the transition probability $T(\mathbf{s}',\mathbf{s}, \mathbf{a}) = \mathbb{P}(S_{t+1}  = \mathbf{s}' | S_t = \mathbf{s}, A_t = \mathbf{a})$, where $S_{t+1}$, $S_t$, and $A_t$ indicate random variables. Additionally, the agent receives the reward $r_t$ according to the reward function $R(\mathbf{s}, \mathbf{a})$.

The agent's behavior is defined by its policy $\pi$, namely the set of rules that the agent exploits to select a specific action in a given state. The policy may be either stochastic $\pi:\mathcal{S} \times\mathcal{A} \to [0,1]$ or deterministic $\pi:\mathcal{S} \to \mathcal{A}$. In the former case, the policy returns the probability of taking an action $\mathbf{a}$ with the environment in state $\mathbf{s}$, that is $\mathbb{P}(\mathbf{a}|\mathbf{s})$. Instead, the latter setting takes into account a deterministic state-to-action map: $\mathbf{a} = \pi(\mathbf{s})$. The agent's goal is to find the optimal policy, that is the policy that maximizes the total expected cumulative reward starting from state $\mathbf{s}$ at time $t$:
\begin{equation}
    V_\pi(\mathbf{s}) = \mathbb{E}_{\pi}\left[G_t|S_t=\mathbf{s}\right] = \mathbb{E}_{\pi}\left[\sum_{k=0}^{N_t} \gamma^k \, r_{t+k}|S_t=\mathbf{s}\right]\, ,
\label{eq:value}
\end{equation}
where $G_t$ denotes the return,  $\gamma \in [0, 1)$ is the discount factor, $\mathbb{E}_{\pi}$ is the expected value of a random variable given that the policy $\pi$ is driving the system evolution, and $N_t$ is the horizon over which actions are taken, which may be either finite or infinite. The total expected cumulative reward $V_{\pi}:\mathcal{S} \rightarrow \mathbb{R}$ is typically referred to as \emph{value function}. The value function quantifies the long-term value of a state when following a fixed policy $\pi$, i.e., how desirable is to be in a state given that the policy $\pi$ drives the decisions. Importantly, it is often convenient to write Equation~\eqref{eq:value} in recursive form, thus obtaining the so-called {\em Bellman equation}
\[
V_\pi(\mathbf{s}) = \mathbb{E}_\pi \left[ r_t + \gamma V_\pi(\mathbf{s'}) | S_t = \mathbf{s} \right].
\]
Note that the optimal value function $V$ associated with the optimal policy is given by
\[
\label{optimal_value_func}
V(\mathbf{s}) = \underset{\pi}{\max}\,V_\pi(\mathbf{s}) = \underset{\pi}{\max}\, \mathbb{E}_\pi \left[ r_t + \gamma V_\pi(\mathbf{s'}) | S_t = \mathbf{s} \right].
\]

\subsection{Twin Delayed Deep Deterministic Policy Gradient}
\label{td3}
In this work, we employ the twin delayed deep deterministic policy gradient (TD3) algorithm~\cite{fujimoto2018addressing}. TD3 is a state-of-the-art off-policy actor-critic algorithm, suitable for continuous control and with high sample efficiency, that learns a deterministic policy $\pi(\mathbf{s};\boldsymbol{\phi})=\pi_{\phi}(\mathbf{s})$ -- the actor -- and two action-value function $Q(\mathbf{s},\mathbf{a};\boldsymbol{\theta}_1)= Q_{\boldsymbol{\theta}_1}(\mathbf{s}, \mathbf{a})$ and $Q(\mathbf{s}, \mathbf{a};\boldsymbol{\theta}_2)= Q_{\boldsymbol{\theta}_2}(\mathbf{s}, \mathbf{a})$ -- the critic(s) -- where $\boldsymbol{\phi}$, $\boldsymbol{\theta}_1$, and $\boldsymbol{\theta}_2$ indicate the parameters of the three neural networks approximating the actor and the two critics, respectively. The actor and the critics are associated with slowly updated target networks 
$\pi_{\bar{\boldsymbol{\phi}}}$, $Q_{\bar{\boldsymbol{\theta}}_{1}}$ and $Q_{\bar{\boldsymbol{\theta}}_{2}}$. Target networks are updated using a soft update rule
\begin{equation}
\begin{split}
   \bar{\boldsymbol{\phi}} &\leftarrow \tau \boldsymbol{\phi} + (1-\tau)\bar{\boldsymbol{\phi}} \\
   \bar{\boldsymbol{\theta}}_{i} &\leftarrow \tau \boldsymbol{\theta}_{i} + (1-\tau)\bar{\boldsymbol{\theta}}_{i}, \quad i=1,2, \\
\end{split}
\end{equation}
with $\tau \ll 1$.
To mitigate overestimation bias, TD3 constructs the target value for critic updates by taking the minimum of the two target critics:
\begin{equation}
y_t = r_t + \gamma \min_{i=1,2} Q_{\bar{\boldsymbol{\theta}}_{i}}\!\left(\mathbf{s}_{t+1}, \pi_{\bar{\boldsymbol{\phi}}}(\mathbf{s}_{t+1})\right),
\end{equation}
where $\pi_{\bar{\phi}}$ denotes the target actor network. Moreover, to solve the exploration problem of deterministic policies during training, we begin with a warm-up phase where actions are chosen randomly; then, we explicitly force action exploration by adding noise to the policy:
\begin{equation}
    \mathbf{a}_t = \pi_{\phi}(\mathbf{s}_t) + \mathbf{\epsilon}
\end{equation}
where the noise is gaussian $\mathbf{\epsilon} \sim \mathcal{N}\left(0;\, \bar{\sigma} \right)$.\\
Each critic is trained by minimizing the Huber loss temporal-difference error, which is suitable for RL in a wide range of parametric settings \cite{botteghi2025hyperl}:
\begin{equation}
\mathcal{L}_{Q_i}(\boldsymbol{\theta}_{i}) =
\mathbb{E}_{(\mathbf{s}_t,\,\mathbf{a}_t,\, r_t,\,\mathbf{s}_{t+1}) \sim \mathcal{B}}
\left[ \, \mathrm{Huber}
\big(Q_{\boldsymbol{\theta}_{i}}(\mathbf{s}_t,\,\mathbf{a}_t) - y_t \big)
\right]\, ,
\end{equation}
where $\mathcal{B}$ indicates the memory buffer collecting the data obtained through the interaction with the environment. 
The actor is updated by minimizing:
\begin{equation}
    \mathcal{L}_{\pi}(\boldsymbol{\phi}) = \mathbb{E}_{\mathbf{s}_t \sim \mathcal{B}}[- Q_{\boldsymbol{\theta}_1}(\mathbf{s}_t, \pi_{\boldsymbol{\phi}}(\mathbf{s}_t))]\, ,
\label{eq:policytd3}
\end{equation}
which gives an update step coherent with the deterministic policy gradient theorem \cite{silver2014deterministic}
\begin{equation}
\nabla_{\boldsymbol{\phi}} \mathcal{L}_{\pi}(\boldsymbol{\phi})
=
-\mathbb{E}_{\mathbf{s}_t \sim \mathcal{B}}
\left[\nabla_{\mathbf{a}} Q_{\boldsymbol{\theta}_{1}}(\mathbf{s}_t,\mathbf{a})\big|_{\mathbf{a}=\pi_{\boldsymbol{\phi}}(\mathbf{s}_t)}
\nabla_{\boldsymbol{\phi}} \pi_{\boldsymbol{\phi}}(\mathbf{s}_t)
\right].
\end{equation}

\section{Methodology}\label{sec:methodology}
\label{methodology}
Towards our goal of learning flow-aware navigation strategies, we consider the navigation problem of a mobile particle in a double gyre flow~\cite{gunnarson2021learning,krishna2023finite,botteghi2025hyperl}. The flow field is defined in a spatial domain $\mathcal{D} \subset \mathbb{R}^2$ as:
\begin{equation*}
    \phi(x, y, t) = A\sin{\big(\pi f(x,t)\big)}\sin{\big(\pi y\big)} \, , 
\end{equation*}
where $x,y$ are the spatial coordinates, $t \in (0,T)$ denotes time, 
and
\begin{equation*}
    f(x,t) = \left( \epsilon \sin(\omega t)\right) x^2 + \left(1 - 2\epsilon \sin(\omega t) \right) x \, . 
\end{equation*}
The velocity field derives from the flow field as $v_x = - \partial\phi / \partial y$ and $v_y = \partial\phi / \partial x$. Hence, the ODE system governing the dynamics of an active particle in the gyre flow field can be written as follows:
\begin{equation}
\left\{
\begin{array}{rlll}
   \displaystyle  \frac{dx}{dt}(t) &= v_x(x(t),\,y(t),\,t) + u_x(t) \\&=  -\pi \, A \,\sin{\big(\pi \, f(x(t),\,t)\big) \, \cos{(\pi \, y(t))} + u_x(t)}\, , & t \in (0,T) \vspace{0.2cm} \\
    \displaystyle  \frac{dy}{dt}(t) &= v_y(x(t),\,y(t),\,t) + u_y(t) \\&= \displaystyle  \pi \, A \, \cos{\big(\pi \,f(x(t),\,t)) \,\sin{(\pi \, y(t)\big)} \, \frac{\partial f}{\partial x}(x(t),\,t) + u_y(t)}\, , & t \in (0,T) \\
    x(0) & = x_0 & & \\
    y(0) & = y_0 & & \\
    \end{array}
    \right.
\label{eq:gyre_flow}
\end{equation}
in which $u_x, u_y$ are the control inputs -- with $\mathbf{a}_t = [u_x(t), u_y(t)]^T \in \mathbb{R}^2$ -- and $\mathbf{x_0} = (x_0, y_0)^T$ is the particle's initial position in the domain.  
The flow is restrained within a domain $(x,y)^T \in \mathcal{D}=(0,2)\times(0, 1)$ and $A, \epsilon, \omega$ are model parameters controlling the flow pattern. In particular, $A$ affects the velocity magnitude, $\epsilon$ the amplitude of the double gyre oscillation in the $x-$direction, and $\omega$ the angular oscillation frequency. Similarly to \cite{krishna2023finite}, we solve~\eqref{eq:gyre_flow} for a fixed time horizon with final time $T=80$s and time step $\Delta t=0.1$s. However, differently from \cite{gunnarson2021learning,krishna2023finite}, where the agent is trained to reach a single target, we consider a more challenging control problem: \emph{navigation and station-keeping to arbitrary targets in a parametric gyre flow}. To allow the agent to learn an advanced navigation strategy rather than a single path, we randomly sample the starting point $\mathbf{x_0}=(x_0, y_0)^T$, the target location $\mathbf{x_g}=(x_g, y_g)^T$, and the parameters of the gyre $\boldsymbol{\mu} = (A, \omega)^T$ for each episode of training and testing. Starting and ending points are chosen uniformly in the subdomain $\mathcal{D'} = [0.1, \,1.9] \times [0.1,\,0.9]$, while gyre parameters are uniformly sampled in the intervals $A \in [0 ,0.5]$, $\omega \in [0.5, 2\pi/3]$, with  $\epsilon$ fixed and equal to $0.1$. As the largest flow velocity is $\approx A\pi$, we force the controller to exploit the dynamics of the gyre flow during navigation by limiting the control action $\mathbf{a}_t = [u_x(t), u_y(t)]^T$ to $u_x, u_y \in [-0.2, 0.2]$. All simulations are performed in a custom-built double-gyre gymnasium environment.

Biological swimmers usually navigate fluids by sensing the local features of the underlying flow, like velocity, vorticity, or pressure. In particular, the work in~\cite{oteiza2017novel} showed that the lateral line of the zebrafish acts as a vorticity sensor. Following this intuition and leveraging the initial results presented in~\cite{gunnarson2021learning}, we solve the aforementioned problem with a bio-inspired approach by constructing agents with different observables:
\begin{itemize}
    \item[(i)] \emph{naive agent} -- the agent can only observe the absolute time frame $t$ and its current relative position with respect to the target $\mathbf{\Delta x} = \mathbf{x}(t) - \mathbf{x_g}$, so its observation vector is constructed as $\mathbf{s}_t = (t,\, \Delta x_t,\, \Delta y_t)^T \in \mathbb{R}^3$.\\
    \item[(ii)] \emph{velocity agent} -- the agent can observe its current relative position with respect to the target and the velocity of the flow field beneath it $v_{i,\,t} = v_i(x(t), \, y(t), \,t)$ with $i=x,y$, so its observation vector is constructed as $\mathbf{s}_t = (\Delta x_t,\, \Delta y_t, \, v_{x,\,t}, \, v_{y,\,t})^T \in \mathbb{R}^4$.\\
    \item[(iii)] \emph{vorticity agent} -- the agent can observe its current relative position with respect to the target and the vorticity of the flow field beneath it $\zeta_t = \zeta(x(t), \,y(t),\,t)$, so its observation vector is constructed as $\mathbf{s}_t = (\Delta x_t,\, \Delta y_t, \, \zeta_t) \in \mathbb{R}^3$. The vorticity is defined as the curl of a given velocity field: $\mathbf{\zeta} = \nabla \times \mathbf{v}$. In 2D, the expression for the vorticity simply becomes $\zeta = \partial v_{y}/\partial x - \partial v_{x}/\partial y$.
\end{itemize}

Fishes and other natural swimmers are known to possess some form of memory~\cite{pouca2021fish}. To further investigate the critical information to navigate the flow efficiently and allow the agent to build a local map of flow structures, we introduce two additional bio-inspired agents:
\begin{itemize}

    \item[(iv)] \emph{velocity history agent} -- the agent can observe its current relative position with respect to the target, the velocity of the flow field beneath it, and can remember its last $M-1$ velocity measurements, leading to the observation vector $\mathbf{s}_t = (\Delta x_t,\, \Delta y_t, \, v_{x,\,t}, \, v_{y,\,t}, \,\dots, \, v_{x,\,{t-M+1}}, \, v_{y,\,{t-M+1}})^T\in \mathbb{R}^{2M+2}$.\\
    \item[(v)] \emph{vorticity history agent} -- the agent can observe its current relative position with respect to the target, the vorticity of the flow field beneath it, and can remember its last $M-1$ vorticity measurements. Its observation vector is constructed as $\mathbf{s}_t = (\Delta x_t,\, \Delta y_t, \, \zeta_t, \, \dots, \, \zeta_{t-M+1})^T \in \mathbb{R}^{M+2}$.
\end{itemize}

It is clear that these last two agents reduce to the previously introduced \emph{velocity} and \emph{vorticity} agents when the memory parameter is set to $M=1$.\\
At each timestep, the agent receives a reward equal to the sum of a distance penalty, a control penalty, and a bonus for reaching the target:
\begin{equation}
\label{reward_func}
    r_t = -c_t^d - c_t^c + b_t\, ,
\end{equation}
where $c_t^d =  \Delta x_t^2 + \Delta y_t^2$ is the squared distance with respect to the target, the control penalty is the rescaled squared norm of the action $c_t^c = \alpha \Vert \mathbf{a} \Vert^2$, and the bonus $b_t= 1$ is received by the agent only when the distance to target is below a certain threshold $|\Delta \mathbf{x}| < \delta$. Since the episodes do not terminate until $T=80\mathrm{s}$, maximizing the expected cumulative reward $V(\mathbf{s}=\mathbf{x_0})$ from Eq.~\eqref{optimal_value_func} and~\eqref{reward_func} leads to navigation policies that aim to reach the target and stay near it while being as energy-efficient as possible. An episode is considered to be successful only if the agent manages to stay near the target for more than $100$ non consecutive time steps $\Delta t$, i.e. more than $1/8$ of the entire episode.

\section{Numerical Results}
\label{numerical_results}
\begin{figure}[h!]
    \centering
    \begin{minipage}{0.49\textwidth}
        \centering
\includegraphics[width=\textwidth]{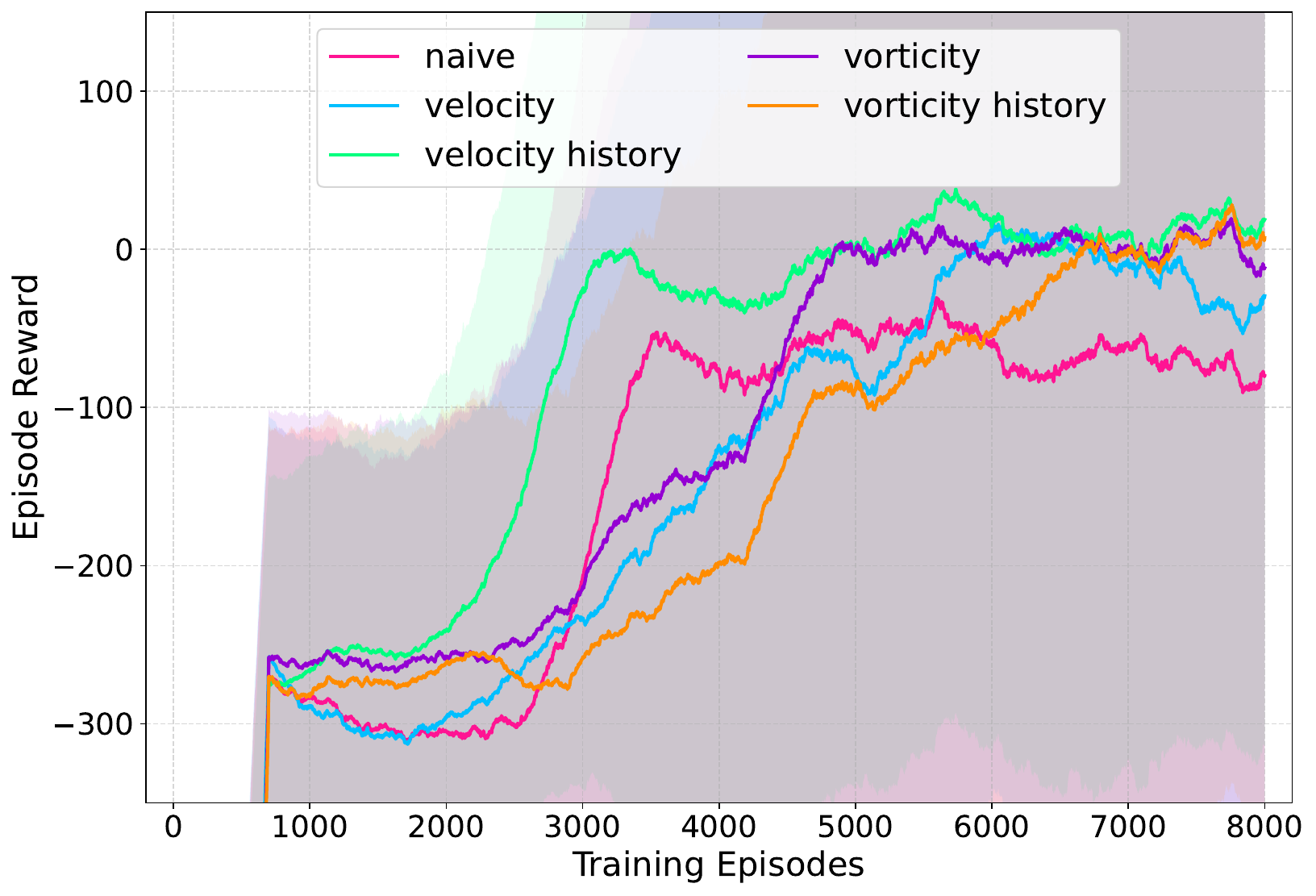}
    \end{minipage}
    \hfill
    \begin{minipage}{0.49\textwidth}
        \centering
    \includegraphics[width=\textwidth]{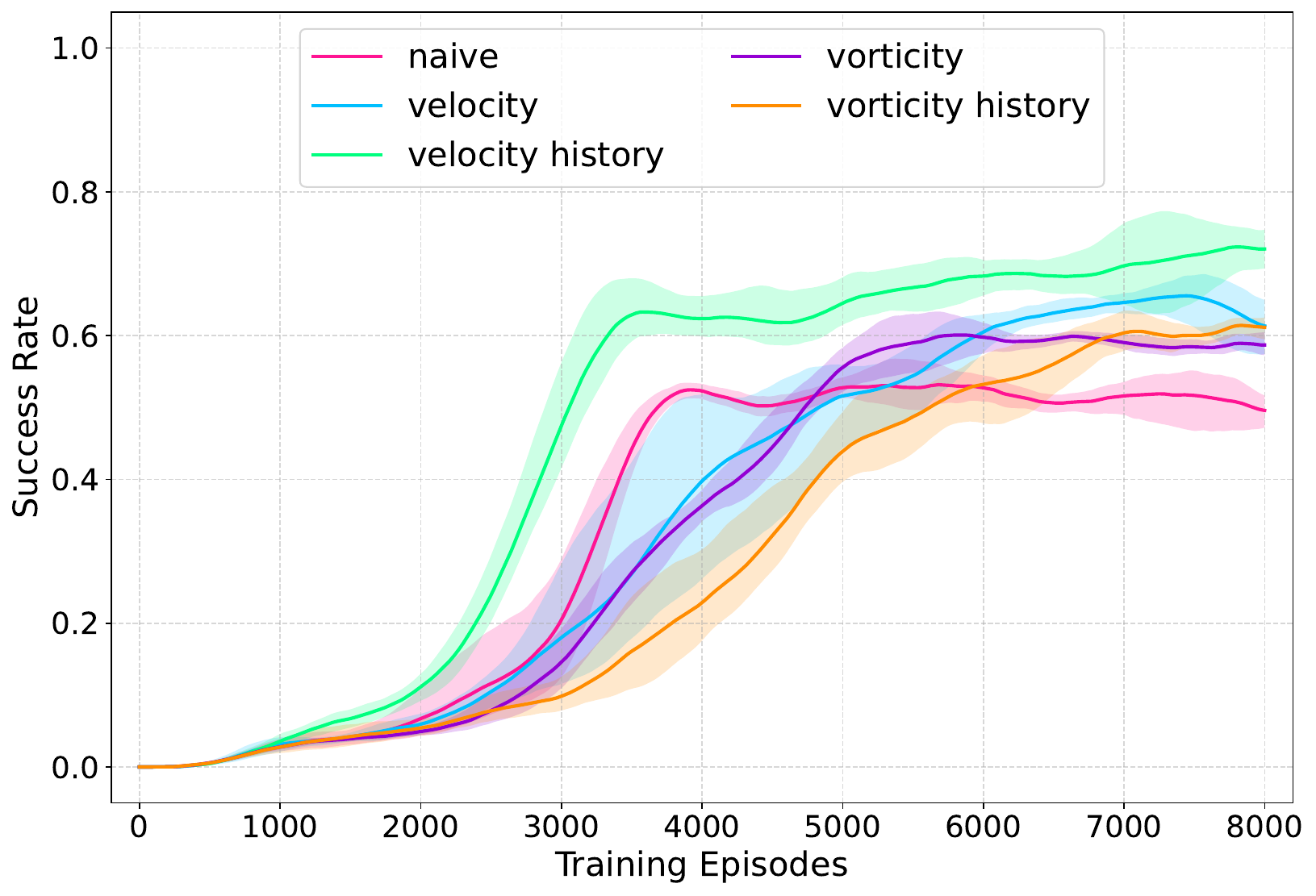}
    \end{minipage}
    \caption{Cumulative training rewards (left panel) and success rates (right panel) for the different agents.}
    \label{reward_sr_blind}
\end{figure}

In the following section, we report the results obtained by training the 5 different agents introduced in Section \ref{sec:methodology}. Each result corresponds the mean of three runs with different seeds (i.e., $1$, $5$, $10$). We set $M=10$ for the \emph{velocity history} and \emph{vorticity history} agents. We motivate and discuss this choice in Appendix \ref{memory_analysis}. We report the complete list of hyperparameters used in our experiments in Appendix \ref{hyperparameters}. Success rate is computed as a moving average on a window of 200 training episodes. Results for the training of the different agents are presented in Fig. \ref{reward_sr_blind}.
\\We can directly remark the following facts:
\begin{itemize}
    \item The \emph{velocity history agent} outperforms every other agent, both in total reward collected and in the success rate ($\sim 71 \%$). Not only it achieves better asymptotic performance, but it also learns faster, achieving peak performance after $\sim 3 \times 10^3$ training episodes. We postulate that this result stems from the ability of the agent to reconstruct the fluid structures accurately.\\
    \item The \emph{naive agent}, probably thanks to its knowledge of the absolute time frame (which is a proxy for the dynamics of the double-gyre), is able to learn a representation of the environment quickly. Nevertheless, its asymptotic performance is the worst, for it never truly learns how to recognize and fully exploit flow features (success rate $\sim 49 \%$).\\
    \item The \emph{vorticity agent} only measures a scalar, while the \emph{velocity agent} measures the two components of the velocity field. However, the former learns faster and is able to achieve performance comparable to the latter (respectively $\sim 58 \%$ and $\sim 60\%$ success rate). This could be due to the intrinsic vorticose properties of the double-gyre environment. Moreover, the vorticity is computed as a spatial derivative of the velocity field, so it is highly informative of flow structures.\\
    \item Finally, the \emph{vorticity history agent} learns slowly but smoothly and has asymptotic performance similar to the \emph{vorticity agent} and \emph{velocity agent}, achieving a success rate of $\sim 60\%$. Remarkably, it eventually converges to a cumulative reward per episode comparable to the \emph{velocity history agent}. For this particular agent, it appears that collecting a history of measures doesn't give an edge in terms of success rate. Notice that the slower learning curve, compared to the \emph{vorticity agent}, can be understood as the increased difficulty of interpreting the variables in a bigger observation vector. 
\end{itemize}


None of the agents manage to achieve $100\%$ success rate, probably due to the difficulty of station-keeping within particularly challenging draws of parameters.

\begin{figure}[t!]
    \centering
    \begin{minipage}{0.49\textwidth}
        \centering
    \includegraphics[width=\textwidth]{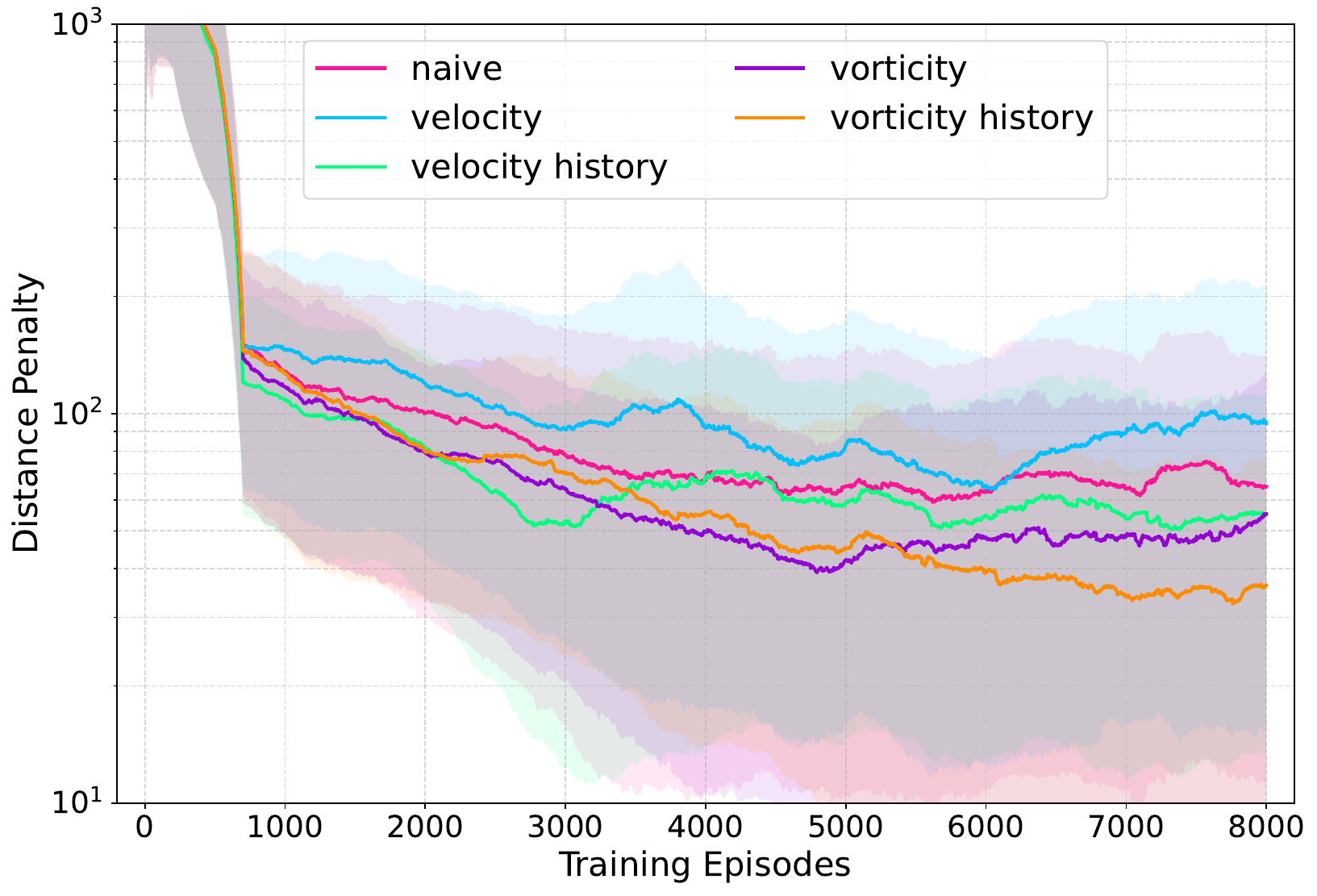}
    \end{minipage}
    \hfill
    \begin{minipage}{0.49\textwidth}
        \centering
    \includegraphics[width=\textwidth]{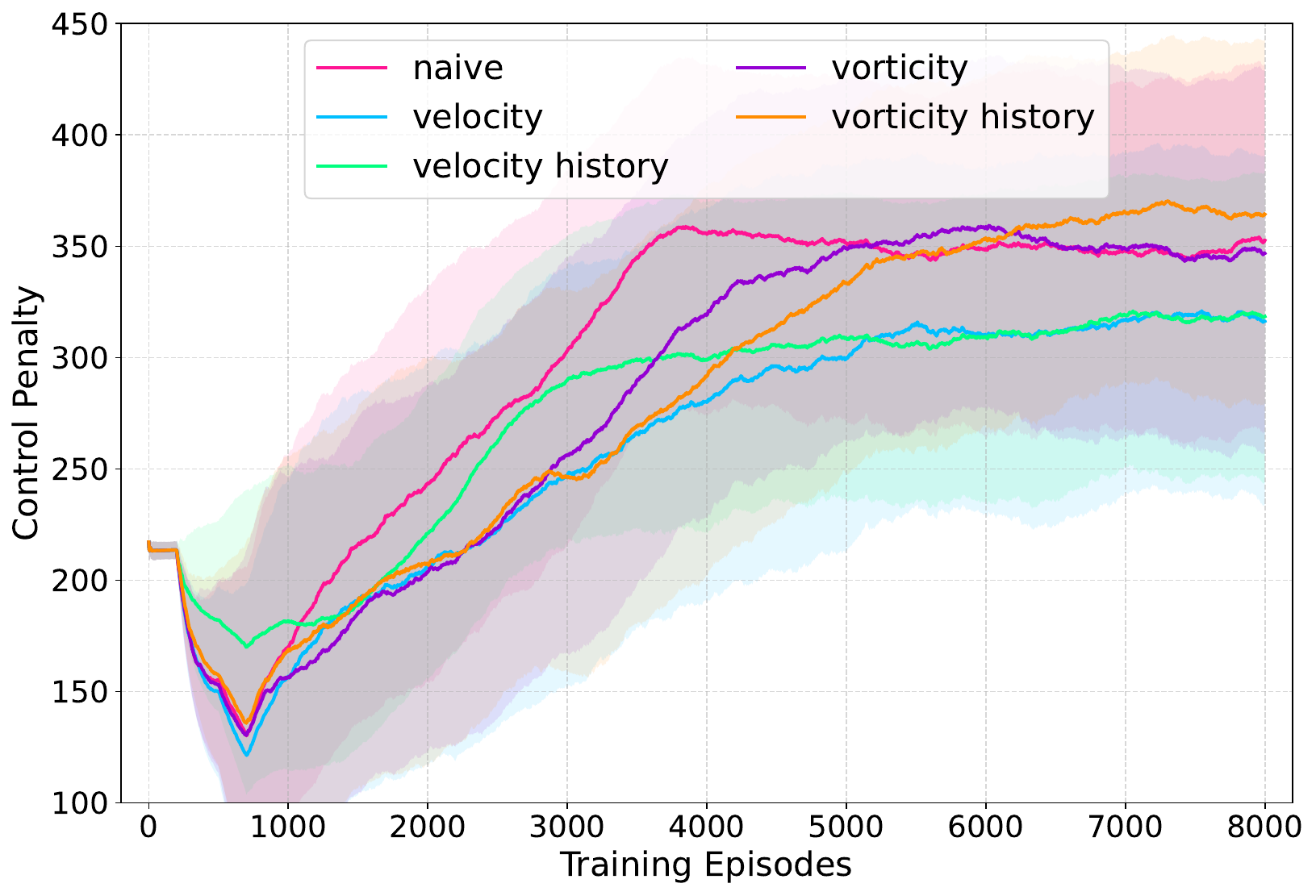}
    \end{minipage}
    \caption{Evolution of cumulative distance penalty (left panel) and control penalty (right panel) during training for the different agents.}
    \label{costs_blind}
\end{figure}

Another interesting set of properties can be deduced by observing the distance penalty and control penalty during training in Fig. \ref{costs_blind}. It appears that different observables induce better optimization of different portions of the reward function. In fact, agents directly measuring the velocity field are able to navigate more energy-efficiently, achieving a remarkably lower cumulative control penalty $C_T^c = \sum_{t=0}^T c_t^c$. On the other hand, vorticity aware agents appear to understand the flow structures and better close in to the target.

\begin{figure}[t!]
    \centering
    \begin{minipage}{0.415\textwidth}
        \centering
    \includegraphics[width=\textwidth]{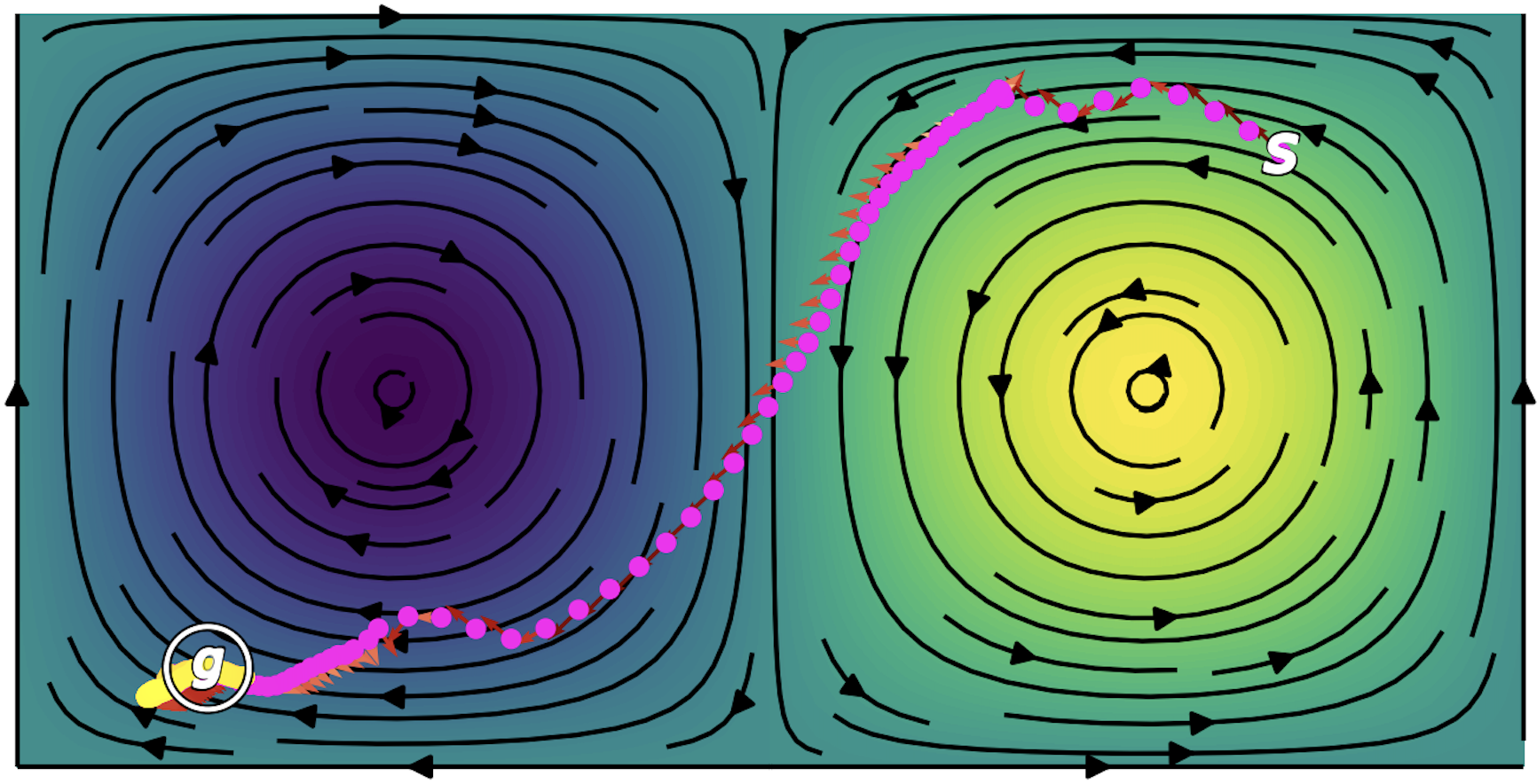}
    \end{minipage}
    \hfill
    \begin{minipage}{0.49\textwidth}
        \centering
    \includegraphics[width=\textwidth]{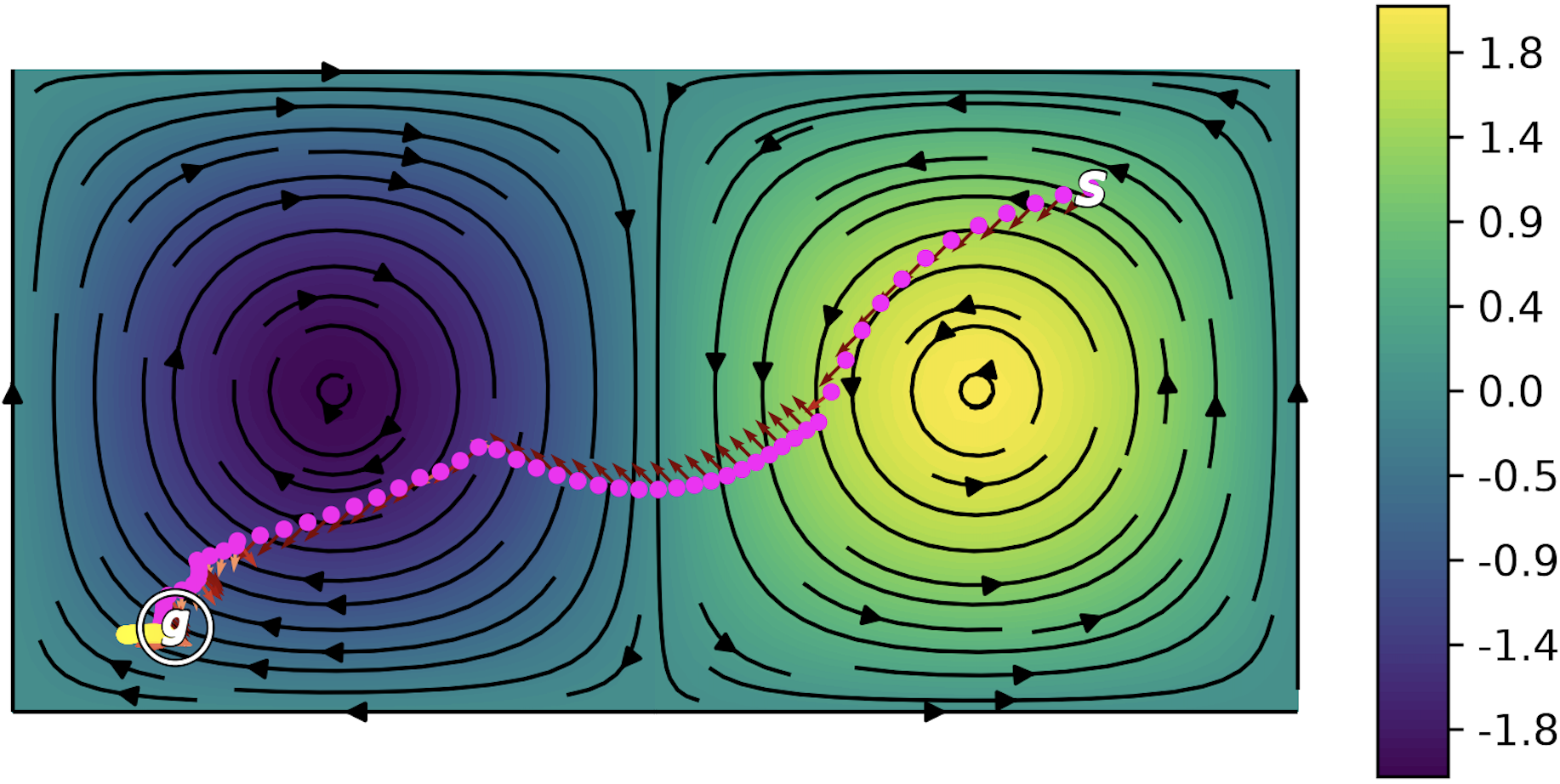}
    \end{minipage}
    \caption{Trajectory controlled by a \emph{velocity history agent} (left panel) and a \emph{naive agent} (right panel). The $\mathbf{S}$ and $\mathbf{g}$ symbols identifies the starting and target position. The colorbar represents the vorticity $\zeta$.}
    \label{trajectories}
\end{figure}

The trajectories in Fig. \ref{trajectories} also show this behavior: even though both agents successfully reach their goal, the \emph{velocity history} one better exploits the flow structures to navigate, and is thus more energy-efficient. These sensor utility insights could be the key to build successful bio-inspired collective leader-follower configurations in swarm settings.



\subsection{Parameter-aware navigation}
\begin{figure}[h!]
    \centering
    \begin{minipage}{0.49\textwidth}
        \centering
    \includegraphics[width=\textwidth]{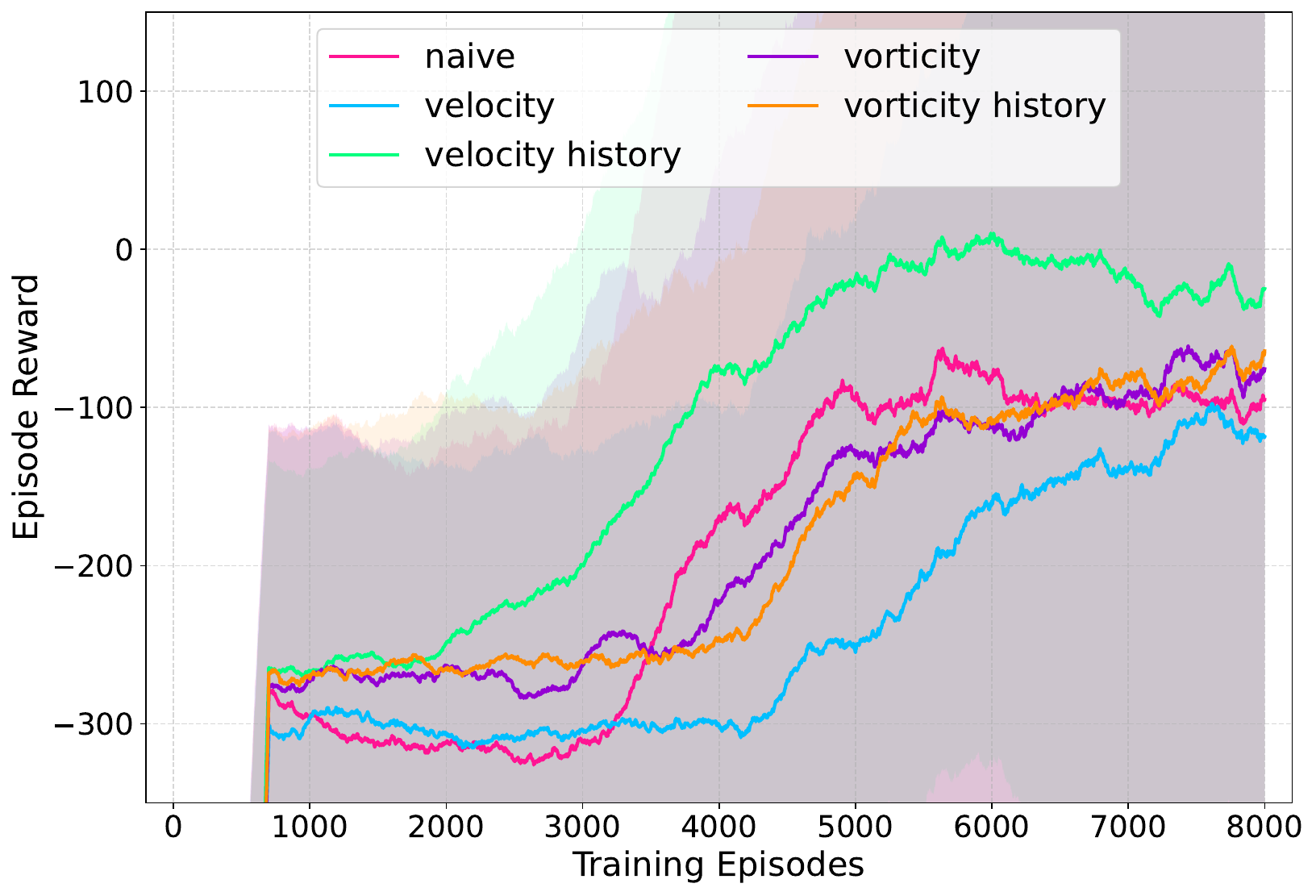}
    \end{minipage}
    \hfill
    \begin{minipage}{0.49\textwidth}
        \centering
    \includegraphics[width=\textwidth]{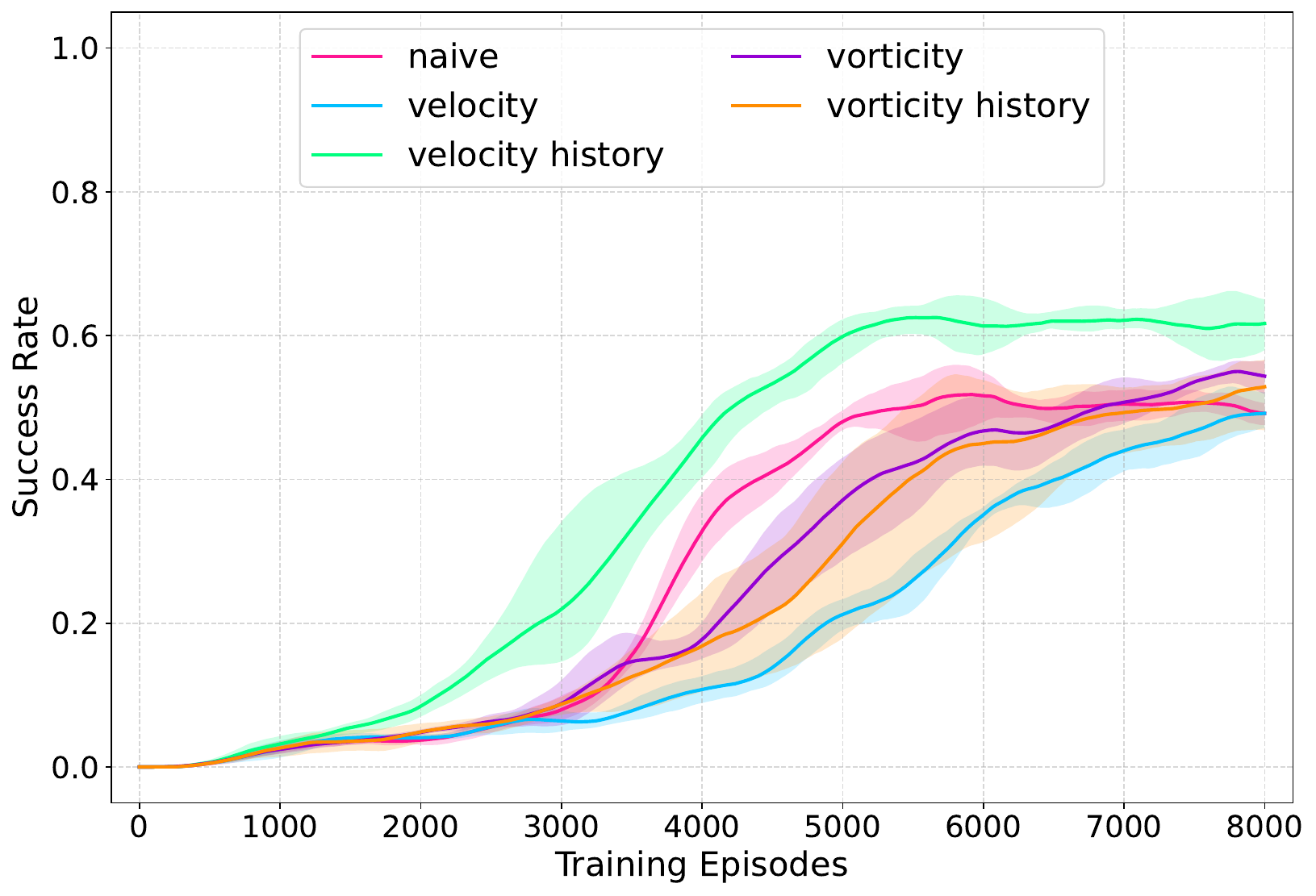}
    \end{minipage}
    \caption{Cumulative training rewards (left panel) and success rate (right panel) for different agents. All agents are informed of the double-gyre parameters $\boldsymbol{\mu} = [A, \omega]$.}
    \label{reward_sr_no-blind}
\end{figure}

To assess whether the performance of the agents are improved by additional knowledge of the environment, we replicate all previous experiments by augmenting the agents observation vector with  the varying parameters of the double-gyre $\boldsymbol{\mu} = (A, \omega)^T \in \mathbb{R}^2$. 
Results are reported in Fig. \ref{reward_sr_no-blind}. Unexpectedly, training rewards and success rate of almost all agents are lower than their parameter-blind counterparts. It appears that agents are unable to infer any meaningful correlation between the local information about the flow and the parameter vector $\boldsymbol{\mu}$ defining a global characterization. 

\begin{figure}[h!]
    \centering
    \begin{minipage}{0.49\textwidth}
        \centering
    \includegraphics[width=\textwidth]{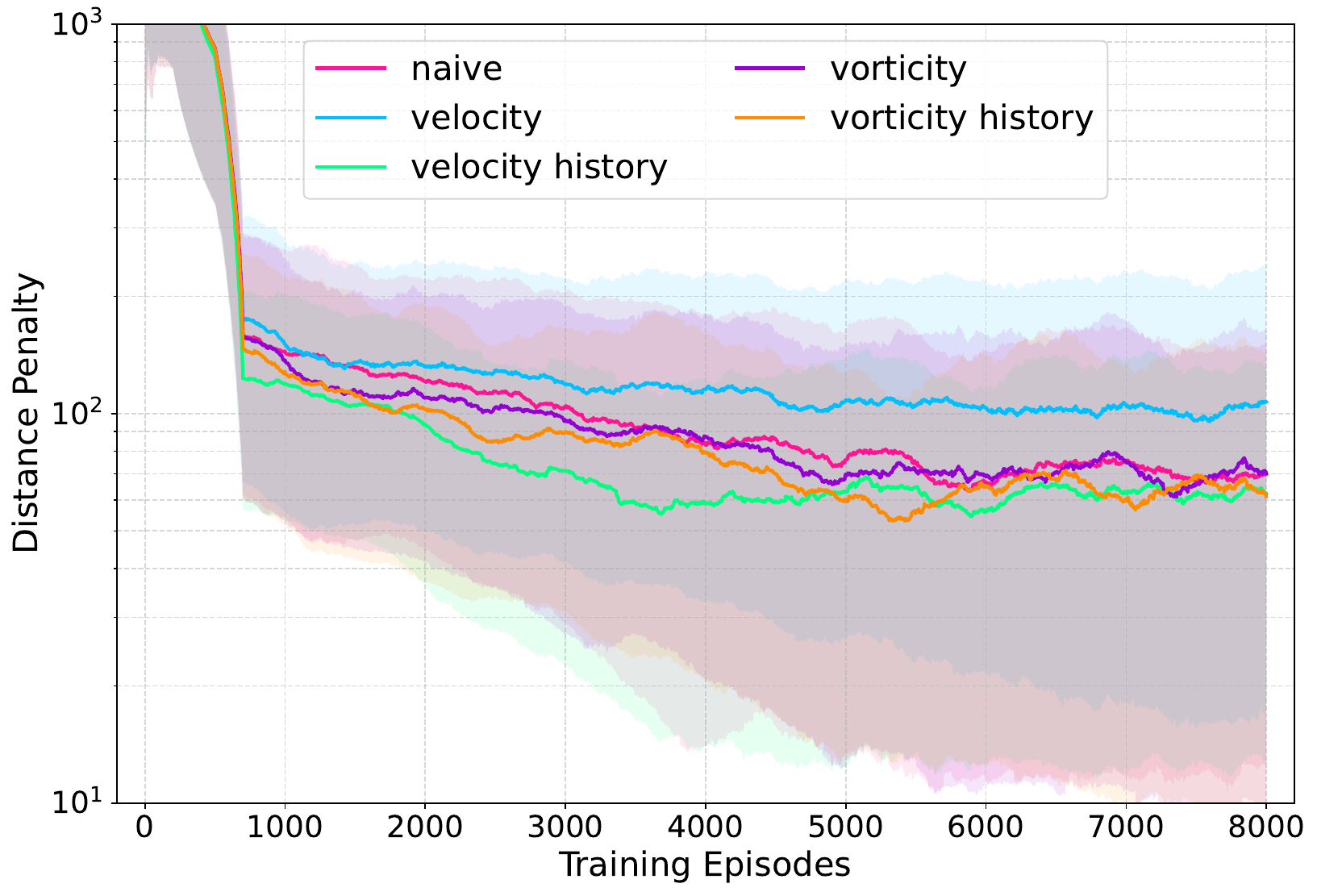}
    \end{minipage}
    \hfill
    \begin{minipage}{0.49\textwidth}
        \centering
    \includegraphics[width=\textwidth]{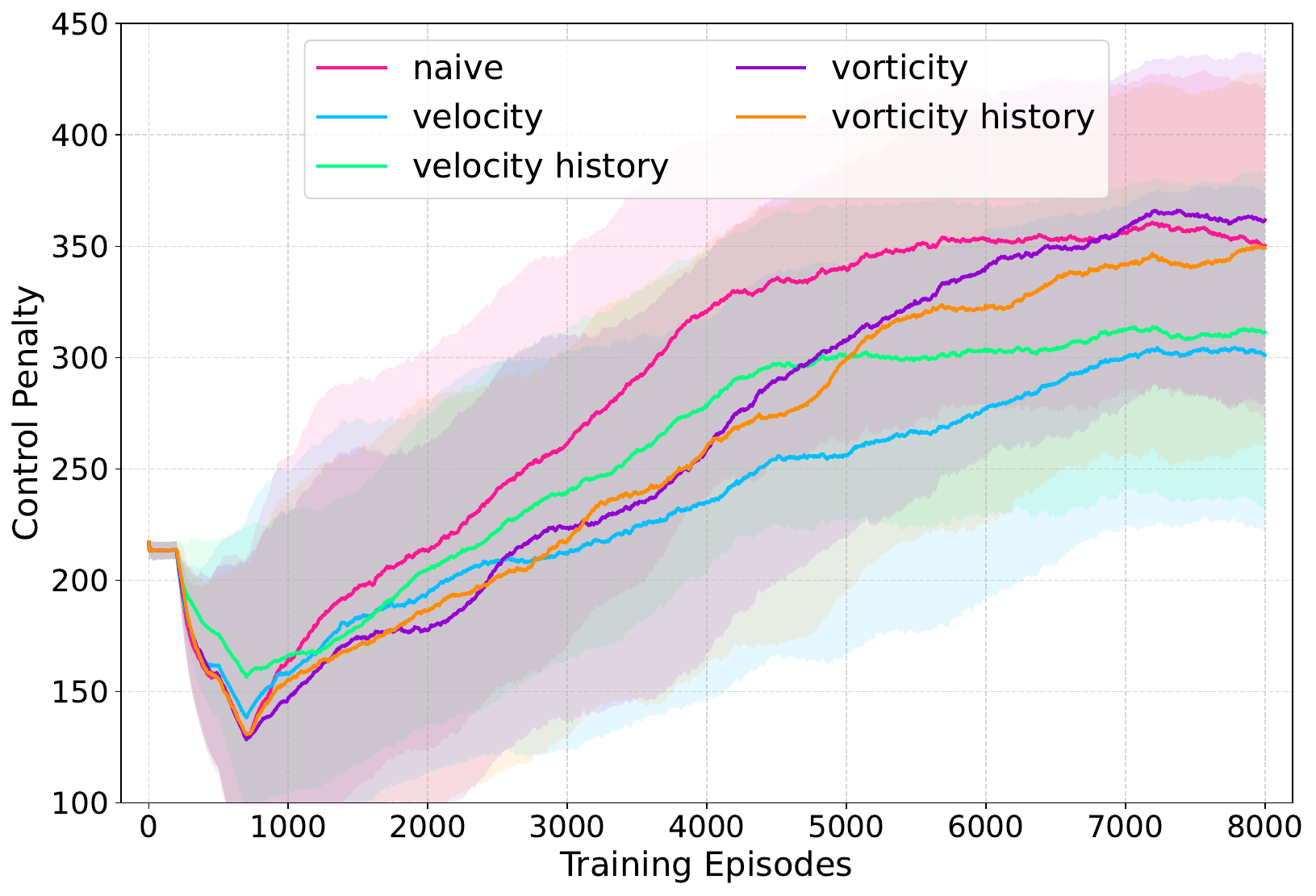}
    \end{minipage}
    \caption{Evolution of cumulative distance penalty (left panel) and control penalty (right panel) during training for the different agents. All agents are informed of the double-gyre parameters $\boldsymbol{\mu} = [A, \omega]$.}
    \label{costs_no-blind}
\end{figure}

The \emph{naive agent} is the only one almost unaffected by the presence of $\boldsymbol{\mu}$ in its observation vector with the success rate still $\sim 49 \%$. The \emph{vorticity agent} performance decrease slightly, achieving a success rate of $\sim 54 \%$. On the other hand, the \emph{velocity} aware agent is deeply affected by the observation of the gyre parameters, as reflected by its success rate that drops to $\sim 49 \%$. Training rewards show that agents with memory are less sensitive to the presence of $\boldsymbol{\mu}$. In fact, even though still asymptotically lower than their "blind" counterparts, \emph{velocity history} and \emph{vorticity history} agents learn smoothly, supporting the claim of their deeper understanding of the environment.\\
Finally, Fig. \ref{costs_no-blind} confirms the optimization peculiarities of different agents described in the previous subsection, with the sole exception of the \emph{velocity history agent}, whose distance penalty $C_T^d$ is lower than the \emph{vorticity agent}. The presence of $\boldsymbol{\mu}$ negatively affects distance penalty and control penalty in the same way as described for the rewards and success rate.


\section{Conclusion}
In this work, we investigated the use of RL for navigating complex parametric flow fields relying on local sensing. The focus of this research was to find optimal bio-inspired navigation strategies that are able to exploit flow features. In fact, agents have low control authority compared to the external field strength. Nevertheless, they must learn how to reach virtually every point on a parametric and chaotic simulation domain, with the minimum possible effort. Leveraging on existing results~\cite{gunnarson2021learning}, we extended the methodology by giving the agents a fixed-length memory, enabling them to remember past measures and we significantly increase the complexity of the task by considering a parametric double gyre flow and arbitrary targets.


The results underline a trade-off in sensor utility: velocity-aware agents minimize control effort, i.e., energy consumption, while vorticity sensing provides a better intrinsic representation of flow structures, leading to more accurate target approaching policies. Equipping agents with global flow parameters, such as knowledge of the double-gyre parameters, proved to deteriorate performance, suggesting that agents learn more resilient and general policies when forced to rely on implicit flow representation rather than explicit, human-readable parameters.

Building on these results, in future work we aim to study navigation strategies in highly chaotic velocity fields and leverage the proposed methodology to improve coordinated behavior in multi-agent systems.

\begin{credits}
\subsubsection{\ackname} 
NB and AM acknowledge the Project “Reduced Order Modeling and Deep Learning for the real-time approximation of PDEs (DREAM)” (Starting Grant No. FIS00003154), funded by the Italian Science Fund (FIS) - Ministero dell'Università e della Ricerca. AM also acknowledges the project “Dipartimento di Eccellenza” 2023-2027 funded by MUR. 

\subsubsection{\discintname}
The authors have no competing interests to declare that are
relevant to the content of this article. 
\end{credits}
%
%
%
\bibliographystyle{splncs04}
\bibliography{references}

\appendix

\section{Memory sensitivity analysis}
\label{memory_analysis}
\begin{figure}[h!]
    \centering
    \begin{minipage}{0.49\textwidth}
    \centering
    \includegraphics[width=\textwidth]{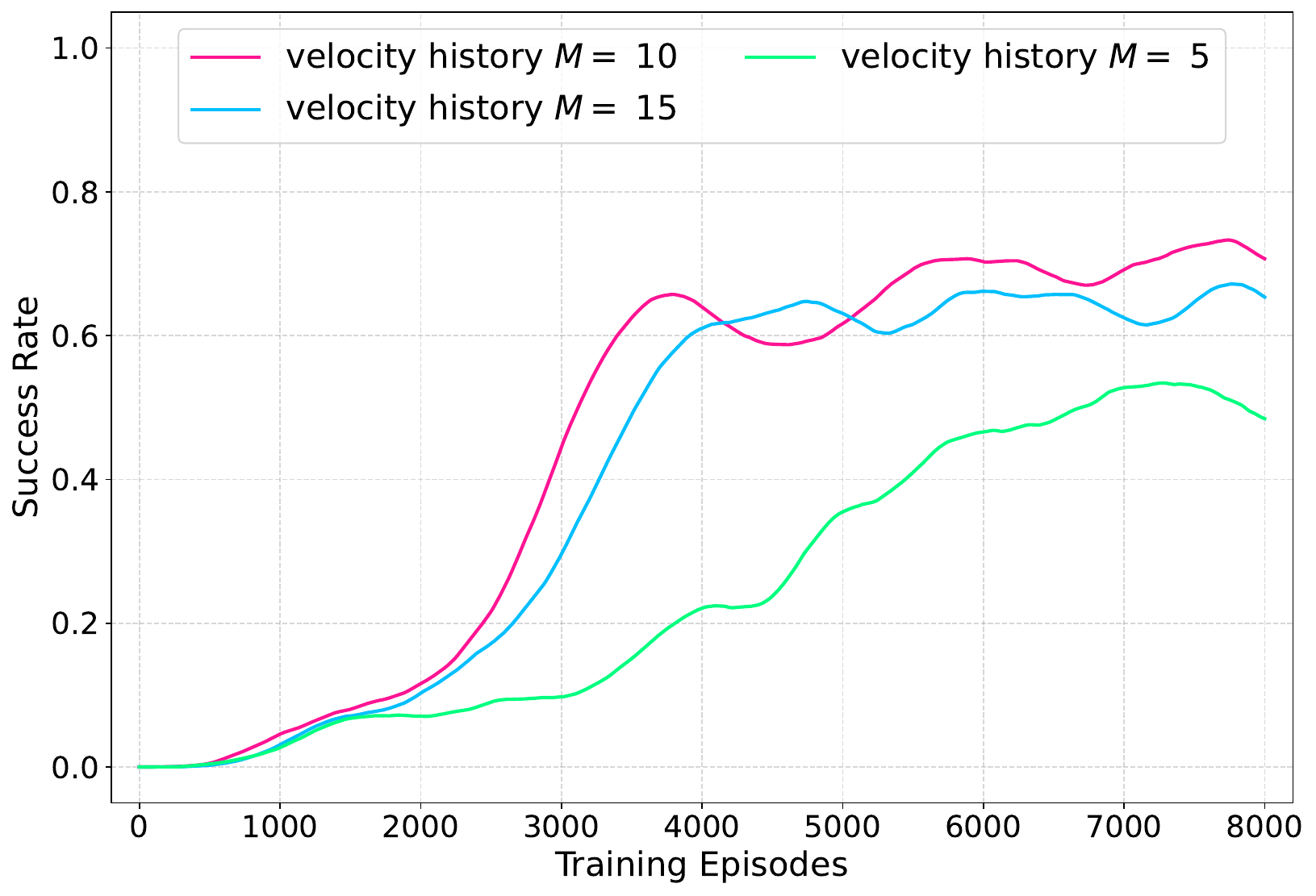}
    \end{minipage}
    \hfill
    \begin{minipage}{0.49\textwidth}
    \centering
    \includegraphics[width=\textwidth]{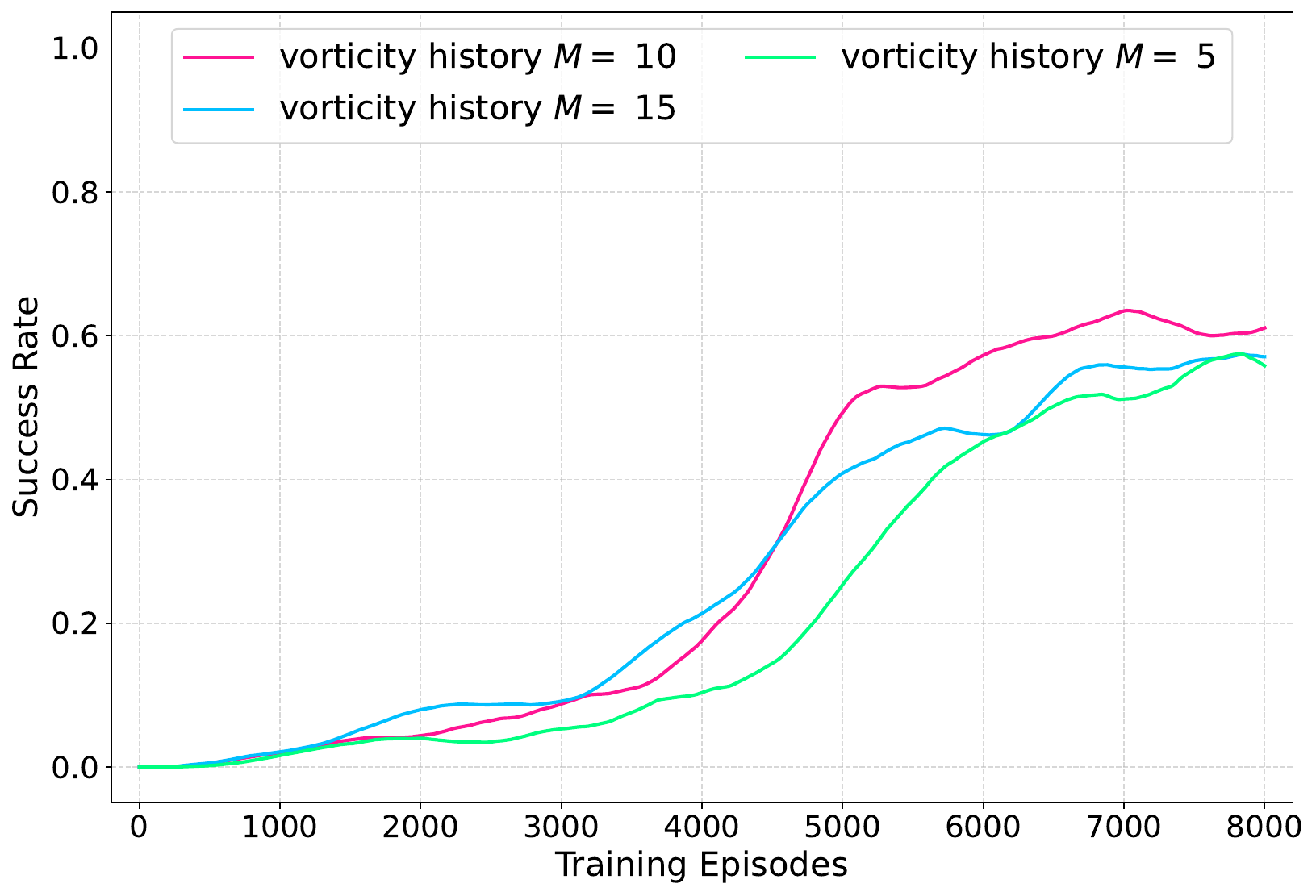}
    \end{minipage}
    \caption{Success rates during training for the \emph{velocity history agent} (left) and the \emph{vorticity history agent} (right) with different values of the memory parameter $M$.}
    \label{success_rate_memory_analysis}
\end{figure}

The memory parameter $M$ affects the amount of information in the \emph{velocity history} and \emph{vorticity history} agents observation vectors. Intuitively, on the one hand longer memory allows an agent to better track the evolution of its surroundings. On the other hand, bigger observation vectors are more difficult to interpret and could provide information on long-gone flow features that are useless for the current situation of the agent. This kind of reasoning calls for a heuristic memory sensitivity analysis of the two agents considered. The results reported in Fig. \ref{success_rate_memory_analysis} are obtained choosing three different values for the memory parameter: $M \in \{5, \,10, \,15\}$. All trainings are done in the "blind" setting, i.e. the agents don't have access to the parameters $\boldsymbol{\mu} = [A, \,\omega]$ of the double-gyre.

Success rates, as well as training rewards (not reported here for the sake of brevity), suggest that both in the \emph{velocity history} and in the \emph{vorticity history} agent an intermediate value of memory $M=10$ is a good choice to achieve the best of flow feature representation and observation clarity.

\section{Hyperparameters of the experiments}
\label{hyperparameters}

\begin{table}[H]
\centering
\label{tab:hyperparameters}
\resizebox{\textwidth}{!}{
\begin{tabular}{ll@{\hspace{2em}}ll@{\hspace{2em}}ll}
\toprule
\multicolumn{2}{l}{\textbf{Network Architecture}} & \multicolumn{2}{l}{\textbf{Optimization}} & \multicolumn{2}{l}{\textbf{Training Configuration}} \\
\cmidrule(r){1-2} \cmidrule(lr){3-4} \cmidrule(l){5-6}
\textbf{Hyperparameter} & \textbf{Value} & \textbf{Hyperparameter} & \textbf{Value} & \textbf{Hyperparameter} & \textbf{Value} \\
\midrule
Actor Architecture & MLP & Learning Rate & $3 \times 10^{-4}$ & Training Episodes & 8,000 \\
Critic Architecture & MLP & Training Batch Size & 256 & Warmup Episodes & 200 \\
Hidden Layers & 2 & Discount Factor ($\gamma$) & 0.99 & Steps per Episode & 801 \\
Hidden Size & 256 & Target Update Rate ($\tau$) & 0.005 & Target Threshold ($\delta$) & 0.055\\
 & & Gaussian Policy Noise ($\bar{\sigma}$) & 0.2 & Control scale ($\alpha$) & 10\\
\bottomrule
\end{tabular}%
}
\end{table}

\end{document}